\documentclass[journal]{IEEEtai}

\usepackage[colorlinks,urlcolor=blue,linkcolor=blue,citecolor=blue]{hyperref}

\usepackage{color,array}

\usepackage{graphicx}

\usepackage{graphicx}%
\usepackage{multirow}%
\usepackage{amsmath,amssymb,amsfonts}%
\usepackage{amsthm}%
\usepackage{mathrsfs}%
\usepackage{xcolor}%
\usepackage{textcomp}%
\usepackage{manyfoot}%
\usepackage{booktabs}%
\usepackage{algorithm}%
\usepackage{algorithmicx}%
\usepackage{algpseudocode}%
\usepackage{listings}%
\usepackage{tabularx}
\usepackage{comment}
\usepackage{mathtools}
\usepackage{xcolor}
\usepackage{subcaption}
\usepackage{pdflscape} 
\usepackage{siunitx}
\usepackage{xspace}
\usepackage{float} 
\usepackage{booktabs} 
\usepackage{siunitx}  
\usepackage{multirow} 
\usepackage{makecell}
\usepackage{rotating}
\usepackage{newpxtext}

\makeatletter
\newcommand{\eg}{\emph{e.g.}\@ifnextchar.{\!\@gobble}{}}
\newcommand{\ie}{\emph{i.e.}\@ifnextchar.{\!\@gobble}{}}

\begin{document}
\title{Constrained Centroid Clustering: A Novel Approach for Compact and Structured Partitioning} 

\author{Sowmini Devi Veeramachaneni, Ramamurthy Garimella 
}
\thanks{Sowmini is with AI \& CS Department, Mahindra University, Hyderabad, India (e-mail: sowminidevi.v@mahindrauniversity.edu.in).}
\thanks{Ramamurthy is with AI \& CS Department, Mahindra University, Hyderabad, India (e-mail: rama.murthy@mahindrauniversity.edu.in).}

\maketitle

\begin{abstract}
This paper presents \emph{Constrained Centroid Clustering (CCC)}, a method that extends classical centroid-based clustering by enforcing a constraint on the maximum distance between the cluster center and the farthest point in the cluster. Using a Lagrangian formulation, we derive a closed-form solution that maintains interpretability while controlling cluster spread. To evaluate CCC, we conduct experiments on synthetic circular data with radial symmetry and uniform angular distribution. Using ring-wise, sector-wise, and joint entropy as evaluation metrics, we show that CCC achieves more compact clusters by reducing radial spread while preserving angular structure, outperforming standard methods such as K-means and GMM. The proposed approach is suitable for applications requiring structured clustering with spread control, including sensor networks, collaborative robotics, and interpretable pattern analysis.
\end{abstract}

\begin{IEEEImpStatement}
Traditional clustering methods often fail to produce compact and interpretable clusters when applied to spatially structured data. This work proposes Constrained Centroid Clustering (CCC), which limits the maximum distance between a cluster center and its farthest point, ensuring tighter, more structured groupings. By preserving angular structure and reducing radial spread, CCC is especially effective for circular or radial datasets. The method is well-suited for applications such as sensor networks, collaborative robotics, and interpretable pattern analysis, where controlling cluster geometry and maintaining structure are essential.
\end{IEEEImpStatement}


\section{Introduction}

\IEEEPARstart{I}n an effort to model biological learning mechanisms, researchers proposed mathematical models of biological neurons. These models, including the McCulloch-Pitts neuron, effectively achieved the classification function using the notions of linear and nonlinear separability. Such abstractions of biological neural networks led to the research area of Artificial Neural Networks (ANNs). These early neural network models operated under the supervised learning paradigm, where the classifier has access to the labels of training patterns during the training phase. The networks learned to classify patterns with sufficiently high accuracy (\eg 98\%), with the testing phase utilized to validate the performance of the trained model on unseen data.
However, researchers soon realized that the assumption of the classifier knowing the class label for every training pattern was restrictive. Many real-world applications involve scenarios where label information is incomplete, unavailable, or expensive to obtain \cite{4787647}. This observation led to the research area of unsupervised learning \cite{10.5555/2380985, Bishop2006}, where the goal is to discover inherent structure within the data without relying on explicit labels.

One of the earliest and most widely studied approaches in unsupervised learning is clustering \cite{10.1145/331499.331504}, which seeks to group similar patterns based on a similarity measure. The pioneering K-means clustering algorithm became foundational, wherein the objective is to minimize the sum of squared distances between patterns and their assigned cluster centers. This algorithm and its variants \cite{1427769} led to explosive research in clustering due to its simplicity and effectiveness across various applications in pattern recognition, data mining, and signal processing.

A well-known insight in clustering theory is that, under the objective of minimizing the sum of squared distances, the optimal cluster center corresponds to the centroid of the data points assigned to the cluster. However, this formulation does not account for additional constraints that may arise in practical clustering applications, such as limiting the distance of the furthest (extremal) pattern from the cluster center to control the spread of the cluster or to enforce spatial or operational constraints in resource-constrained systems.
In recent years, constrained clustering has received increased attention, with work exploring the use of instance-level and structural constraints within clustering frameworks to improve stability and interpretability \cite{10.1007/11564126_11,5206852}. Additionally, advances in clustering under deep learning frameworks \cite{8412085} and the development of coreset-based methods for scalable clustering \cite{bachem2017practicalcoresetconstructionsmachine} reflect the need for clustering methods that maintain structure while controlling computational and spatial resources.

Motivated by these practical considerations, the authors of \cite{bradley2000constrained} sought to formulate and solve an ‘optimal’ clustering problem as a constrained optimization \cite{bradley2000constrained} problem in which the extreme pattern belonging to the cluster is required to lie within a fixed distance from the cluster center. Recognizing that in real-world scenarios this distance constraint typically appears as an inequality constraint, this research paper presents an effort to solve such constrained optimization problems systematically within the clustering framework. We refer to the proposed method as \emph{Constrained Centroid Clustering (CCC)}, which extends classical centroid-based clustering by explicitly incorporating a spread-control constraint to limit the distance between the cluster center and the extremal pattern within each cluster while preserving computational simplicity and interpretability.

In addition to this formulation, the paper explores various innovative ideas related to clustering, including the use of probability mass functions (PMFs) over ring-sector partitioning of the data space and the application of entropy to compare and analyze different clustering methods structurally. Information-theoretic measures \cite{JMLR:v11:vinh10a,MEILA2007873} have proven valuable in systematically evaluating clustering methods, and their incorporation in our work enables quantitative analysis of structure and compactness within the proposed CCC framework.

While real-world datasets exhibiting perfect radial symmetry are uncommon, the practical need to control the spread of clusters arises in various applications, including sensor network formation, collaborative robotics, and resource-aware data grouping. Classical clustering methods do not enforce such spread constraints, which can result in clusters with unbounded or undesirable spatial extents in these contexts. The proposed CCC approach systematically incorporates a spread-control constraint within the clustering framework while retaining the interpretability and computational simplicity of centroid-based methods, making it suitable for practical scenarios where bounded, compact clusters are desired.

This research paper is organized as follows. In Section \ref{related}, related research literature on clustering and constrained optimization approaches is reviewed. In Section \ref{constrained}, the constrained clustering-based optimization problem is formulated and solved using the Lagrangian approach. In Section \ref{methods}, several innovative ideas related to clustering methods, including ring-sector clustering and information-theoretic measures, are discussed. Experimental results and their analysis are discussed in Secion \ref{expts}, and finally, the research paper concludes in Section \ref{conclusion} with insights and potential directions for further exploration.
\section{Review of Related Literature}\label{related}
The K-means clustering algorithm, first introduced by McQueen \cite{mcqueen1967some}, has been a widely used method in machine learning for its simplicity and effectiveness in finding patterns in data without labels. By dividing a dataset into $k$ clusters and minimizing the sum of squared distances between each point and its cluster center, K-means provides a practical way to group similar data in an unsupervised setting.

Later, researchers studied K-means from an optimization point of view and showed that, when using the squared Euclidean distance, the best cluster center for a group of points is simply the centroid of those points. This understanding helped explain why the K-means update step works and gave a stronger theoretical basis for its use in practice.

However, the standard K-means algorithm does not handle noise or outliers well, as these can shift the cluster centers and affect the clustering quality. To address this, some researchers explored constrained versions of clustering that add conditions to reduce the impact of noise or to use additional information in the clustering process \cite{wagstaff2001constrained,10.1007/11564126_11,5206852}. These constraints help in obtaining clusters that are more stable and meaningful, especially when the data has noise or when partial supervision or background knowledge is available.

Another related idea is considering extremal patterns, which are points that lie far from the main group within a cluster. In many real-world problems, it is useful to set a limit on how far such points can be from the cluster center, either due to physical constraints or to keep the clusters interpretable \cite{ester1996density}.

In addition to handling constraints, researchers have explored scalable and robust clustering using coresets \cite{bachem2017practicalcoresetconstructionsmachine} and examined the use of deep learning frameworks for clustering to improve representational power while managing cluster structures \cite{8412085}. These developments reflect the broader need for methods that preserve structure and manage computational resources during clustering.

Beyond constrained clustering, researchers have also looked at information-theoretic approaches to study and compare clustering results. Measures like entropy and Kullback-Leibler divergence can help analyze how patterns are distributed within clusters and compare different clustering methods \cite{lin2002divergence, zhong2003unified, JMLR:v11:vinh10a, MEILA2007873}. In this work, we also use these ideas by defining probability mass functions over radial and angular divisions of the data and then using entropy to systematically study and interpret the clustering outcomes.

While constrained clustering has been studied earlier, most of the work focuses on adding pairwise constraints or using partial supervision to guide clustering. There has been less focus on explicitly formulating and solving a clustering problem where the farthest point from the cluster center is kept within a set limit, particularly within a K-means framework. Moreover, finding a closed-form analytical solution using Lagrangian methods and KKT conditions under such a constraint is not commonly discussed in existing work. By addressing this specific case and pairing it with a structured analysis using radial and angular distributions with entropy, 
our work aims to add a practical and interpretable extension to classical clustering methods.

Building on these insights and limitations in existing approaches, we next present our formulation of constrained clustering in Section \ref{constrained}, detailing the problem setup, analytical solution, and interpretation.

\section{Constrained Clustering Approaches}\label{constrained}

\subsection*{Formulation of the Optimization Problem}
To understand constrained clustering, we first review the unconstrained clustering formulation as a baseline. We then introduce additional constraints to control the cluster's spread or position relative to specific reference points, leading to the constrained formulation.

Consider a set of \(N\) pattern vectors in \(d\)-dimensional space:
\[
\{\overline{Y}_1, \overline{Y}_2, \ldots, \overline{Y}_N\},
\]
where:
\[
\overline{Y}_i = (y_{i1}, y_{i2}, \ldots, y_{id}), \quad 1 \leq i \leq N.
\]
We wish to find the cluster center
\[
\overline{Y}_0 = (y_{01}, y_{02}, \ldots, y_{0d})
\]
that minimizes the total squared Euclidean distance to the given \(N\) pattern vectors:
\[
J(\overline{Y}_0) = \sum_{i=1}^{N} \sum_{j=1}^{d} (y_{ij} - y_{0j})^2.
\]
In the unconstrained case, the minimizer is the \emph{centroid}:
\[
y_{0j} = \frac{1}{N} \sum_{i=1}^{N} y_{ij},
\quad j = 1, 2, \ldots, d.
\]
However, in practical scenarios, it may be necessary to \emph{limit the distance between the cluster center and the farthest point in the cluster (the extremal pattern)} to control the spread of the cluster. We define the extremal pattern as the point in the cluster farthest from the centroid of the other patterns.

To formalize this, we introduce a \emph{constraint} that the squared distance between the cluster center and the extremal pattern should not exceed a threshold $S$. Since this extremal pattern is now handled via the constraint, we exclude it from the objective, and thus the minimization is performed over the remaining $N-1$ points.
\[
\begin{aligned}
\min_{\overline{Y}_0} \quad & J(\overline{Y}_0) = \sum_{i=1}^{N-1} \sum_{j=1}^{d} (y_{ij} - y_{0j})^2 \\
\text{subject to} \quad & \sum_{j=1}^{d} (y_{0j} - y_{Nj})^2 \leq S,
\end{aligned}
\]
where \(\overline{Y}_N\) is the extremal pattern.

\subsection*{Lagrangian Formulation}

Let \(\lambda \geq 0\) be the Lagrange multiplier. We define:
\[
\mathcal{L}(\overline{Y}_0, \lambda) = \sum_{i=1}^{N-1} \sum_{j=1}^{d} (y_{ij} - y_{0j})^2 + \lambda \left( \sum_{j=1}^{d} (y_{0j} - y_{Nj})^2 - S \right).
\]

\subsection*{KKT Conditions}

The Karush-Kuhn-Tucker (KKT) conditions for this problem are:
\begin{itemize}
    \item \textbf{Stationarity:}
    \[
    \frac{\partial \mathcal{L}}{\partial y_{0j}} = -2 \sum_{i=1}^{N-1} (y_{ij} - y_{0j}) + 2\lambda (y_{0j} - y_{Nj}) = 0,
    \]
    which simplifies to:
    \[
    y_{0j} = \frac{ \sum_{i=1}^{N-1} y_{ij} + \lambda y_{Nj} }{ N - 1 + \lambda }.
    \]

    \item \textbf{Primal feasibility:} \(\sum_{j=1}^{d} (y_{0j} - y_{Nj})^2 \leq S\).

    \item \textbf{Dual feasibility:} \(\lambda \geq 0\).

    \item \textbf{Complementary slackness:}
    \[
    \lambda \left( \sum_{j=1}^{d} (y_{0j} - y_{Nj})^2 - S \right) = 0.
    \]
\end{itemize}

Here, the Lagrange multiplier \(\lambda\) quantifies how strongly the distance constraint influences the computed cluster center. When \(\lambda = 0\), the constraint is inactive, and the centroid reduces to the unconstrained mean of the points. As \(\lambda\) increases, the cluster center is adjusted to respect the distance constraint imposed by \(S\).

\subsection*{Solution Cases}

\textbf{Case 1:} If the constraint is inactive (\(\lambda = 0\)), then:
\[
y_{0j} = \frac{1}{N - 1} \sum_{i=1}^{N-1} y_{ij},
\]
and we check whether:
\[
\sum_{j=1}^{d} (y_{0j} - y_{Nj})^2 \leq S.
\]
If satisfied, this is the optimal solution.

\medskip

\textbf{Case 2:} If the constraint is active (\(\lambda > 0\)):
\[
y_{0j} = \frac{ \sum_{i=1}^{N-1} y_{ij} + \lambda y_{Nj} }{ N - 1 + \lambda }.
\]
Let:
\[
A_j = \sum_{i=1}^{N-1} y_{ij},
\]
so:
\[
y_{0j} - y_{Nj} = \frac{ A_j - (N - 1) y_{Nj} }{ N - 1 + \lambda }.
\]
Using the constraint:
\[
\sum_{j=1}^{d} \left( \frac{ A_j - (N - 1) y_{Nj} }{ N - 1 + \lambda } \right)^2 = S,
\]
Define:
\[
C = \sum_{j=1}^d \left( A_j - (N-1)y_{Nj} \right)^2
\]
Then:
\[
\frac{C}{(N - 1 + \lambda)^2} = S \Rightarrow
\lambda = \sqrt{ \frac{C}{S} } - (N - 1)
\]

\[
\lambda = \max\left(0, \sqrt{ \frac{ \sum_{j=1}^d \left( \sum_{i=1}^{N-1} y_{ij} - (N-1)y_{Nj} \right)^2 }{ S } } - (N - 1) \right)
\]

Once \(\lambda\) is found, the optimal centroid is:
\[
y_{0j} = \frac{ \sum_{i=1}^{N-1} y_{ij} + \lambda y_{Nj} }{ N - 1 + \lambda }, \quad \forall \quad j = 1, \ldots, d.
\]

\medskip

It is worth noting that as the threshold \(S\) becomes large, the constraint becomes inactive, and the solution reduces to the standard unconstrained centroid, ensuring consistency with classical clustering behavior. The proposed formulation requires only simple summations and the computation of a scalar multiplier \(\lambda\), making it computationally efficient and scalable even in high-dimensional settings.
\paragraph{Practical Implementation Note}
In practice, once the constrained centroids are computed, each point is reassigned to its nearest constrained centroid. To enforce the maximum distance constraint, if a point lies at a distance exceeding \(\sqrt{S}\) from its assigned constrained centroid, it is moved inward along the line connecting it to the centroid so that its distance equals \(\sqrt{S}\). This preserves the angular direction while reducing the radial spread, ensuring practical enforcement of the constraint and directly enabling quantitative evaluation using entropy-based compactness measures.

This constrained clustering approach enables the computation of a cluster center that is representative of the data while ensuring the extremal pattern remains within a specified distance from the cluster center. It is particularly useful in applications requiring bounded cluster spread, such as sensor networks, collaborative robotics, and interpretable, resource-aware clustering in machine learning systems.

\begin{figure}[h!]
    \centering
    \includegraphics[width=1\linewidth]{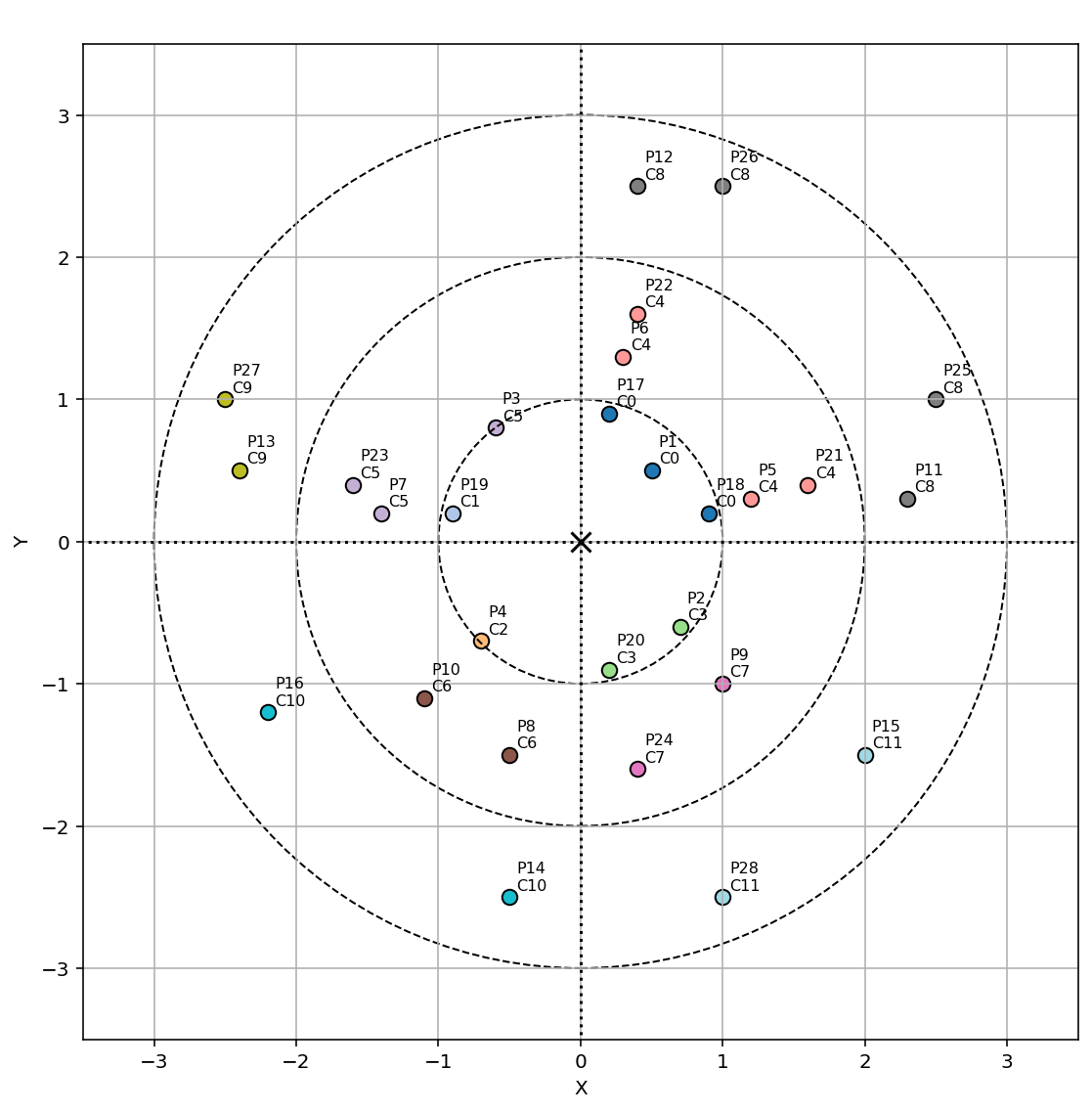}
    \caption{Illustration of ring-sector clustering: data points (P1–P28) grouped based on radial distance and angular position, enabling structured, interpretable clustering with labeled cluster IDs (C0–C11).}
    \label{fig:ring-sector}
\end{figure}
\section{Clustering Methods: Innovative Ideas}\label{methods}
When clustering patterns in a \(d\)-dimensional Euclidean space, the patterns belonging to a cluster typically lie within a hypersphere or a convex region, especially if the data lies within a bounded region. To systematically analyze the structure within a cluster, we associate a \emph{Probability Mass Function (PMF)} with the distribution of patterns lying within concentric hyperspheres of increasing radius from the cluster center.

Let the concentric hyperspheres from the center be indexed by \(\{1, 2, \ldots, R\}\), with \(n_i\) denoting the cumulative number of patterns within the \(i^{th}\) hypersphere and \(n_0 = 0\). We define:
\[
\tilde{n}_i = n_i - n_{i-1},
\]
representing the number of patterns specifically within the \(i^{th}\) ring (excluding inner rings). The ring-based PMF is then computed as:
\[
p_i = \frac{\tilde{n}_i}{\sum_{j=1}^R \tilde{n}_j},
\quad 1 \leq i \leq R.
\]

To incorporate angular distribution into the analysis, we propose a joint PMF that captures both radial and angular characteristics of the clusters. We divide the hyperspheres into angular sectors, resulting in a ring-sector partitioning of the data space. This framework can be visualized in Figure~\ref{fig:ring-sector}, where concentric rings and angular sectors partition the two-dimensional space, assigning structured labels based on radial distance and angular position.

To illustrate the computation process, consider point P1 with coordinates \((0.5, 0.5)\). Its radial distance from the origin is:
\[
r = \sqrt{0.5^2 + 0.5^2} = 0.71,
\]
and its angular position:
\[
\theta = \text{atan2}(0.5, 0.5) \times \frac{180}{\pi} = 45^{\circ}.
\]
Similarly, for other example points: P2 at \((-0.6, 0.7)\) yields \(r = 0.92, \theta = 320^{\circ}\); P7 at \((0.2, -1.4)\) yields \(r = 1.41, \theta = 171.87^{\circ}\); and P10 at \((-1.1, -1.1)\) yields \(r = 1.56, \theta = 225^{\circ}\). These calculations demonstrate how each point's polar coordinates determine its assignment to a specific ring (via \(r\)) and sector (via \(\theta\)).

Let the sectors be indexed by \(\{1, 2, \ldots, J\}\). The joint PMF capturing the distribution across rings and sectors is defined as:
\[
p_{i,j} = \frac{\text{\# of patterns in the } i^{th} \text{ ring and } j^{th} \text{ sector}}{\text{Total number of patterns in the cluster}},
\]
for \(1 \leq i \leq R, 1 \leq j \leq J\).

Additionally, the angular PMF can be computed as:
\[
S_k = \frac{\Gamma_k}{\sum_{i=1}^J \Gamma_i},
\]
where \(\Gamma_k\) is the number of patterns in the \(k^{th}\) sector of the cluster space.

Following principles from information theory, entropy measures associated with these distributions can be computed to quantitatively assess the spread and structure within clusters:
\[
H(P) = -\sum_{i} p_i \log p_i,
\quad
H(S) = -\sum_{k} S_k \log S_k,
\]
\[
\quad
H(P_{i,j}) = -\sum_{i} \sum_{j} p_{i,j} \log p_{i,j},
\]
providing a systematic approach for evaluating and comparing clustering methods under structured, interpretable frameworks. The detailed step-by-step computation of these entropy measures, using the experimental distributions from the example shown in Figure~\ref{fig:ring-sector}, is provided in the Appendix for reference.
\section{Experimental Results}\label{expts}
To evaluate the effectiveness of the proposed constrained clustering method, we conducted a systematic set of experiments on synthetically generated circular data exhibiting radial symmetry and uniform angular distribution. Specifically, we generated a dataset comprising 500 points sampled from an isotropic Gaussian distribution centered at the origin, resulting in a circular spread of points around the center. We performed experiments with different standard deviations of 1.0, 1.2, and 1.5 to examine the method’s performance under varying data spreads. This controlled data structure is particularly suitable for testing clustering algorithms in a scenario where both radial and angular distributions can be explicitly analyzed.

Using this dataset, we applied standard clustering baselines, including KMeans, Gaussian Mixture Models (GMM), DBSCAN, and Agglomerative Clustering, to establish comparative references for evaluation. For each method, we computed ring-wise, sector-wise, and joint entropy measures on the resulting clusters to quantitatively assess the compactness, spread, and structural consistency of the clusters formed under different algorithms.

Ring-wise entropy captures the radial distribution of data points within the clusters, while sector-wise entropy quantifies the angular spread, and the joint entropy reflects the combined spatial distribution in polar coordinates. By employing these metrics, we systematically analyzed how each clustering method partitions the data in terms of radial compactness and angular structure.
\begin{figure*}[t]
    \centering
    \resizebox{1\textwidth}{!}{ 
    \begin{minipage}{\textwidth}
    \begin{subfigure}[b]{0.32\textwidth}
        \includegraphics[width=\textwidth]{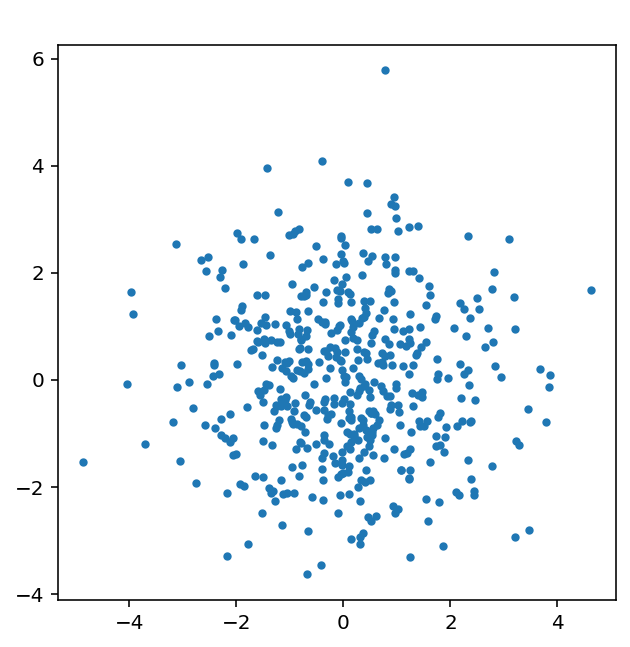}
        \caption{Original Data}
        \label{fig:original}
    \end{subfigure}
    \hspace{0.01\textwidth}
    \begin{subfigure}[b]{0.32\textwidth}
        \includegraphics[width=\textwidth]{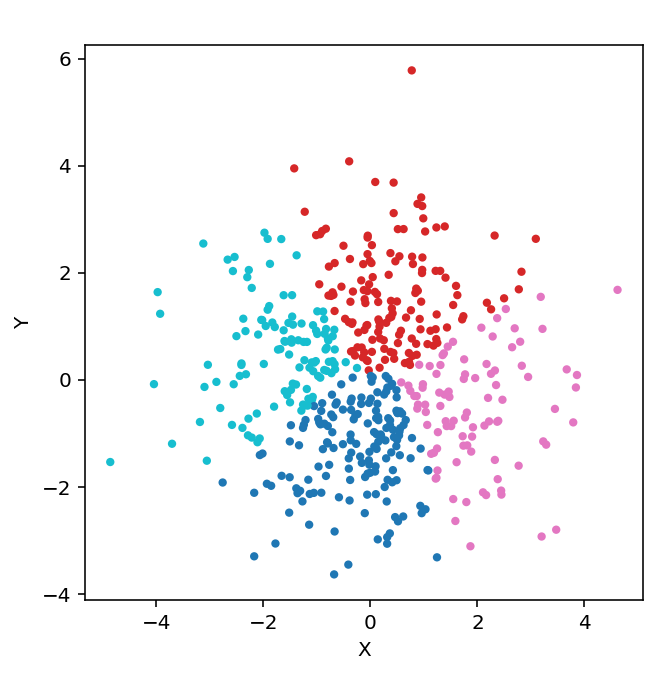}
        \caption{KMeans}
        \label{fig:kmeans}
    \end{subfigure}
    \hspace{0.01\textwidth}
    \begin{subfigure}[b]{0.32\textwidth}
        \includegraphics[width=\textwidth]{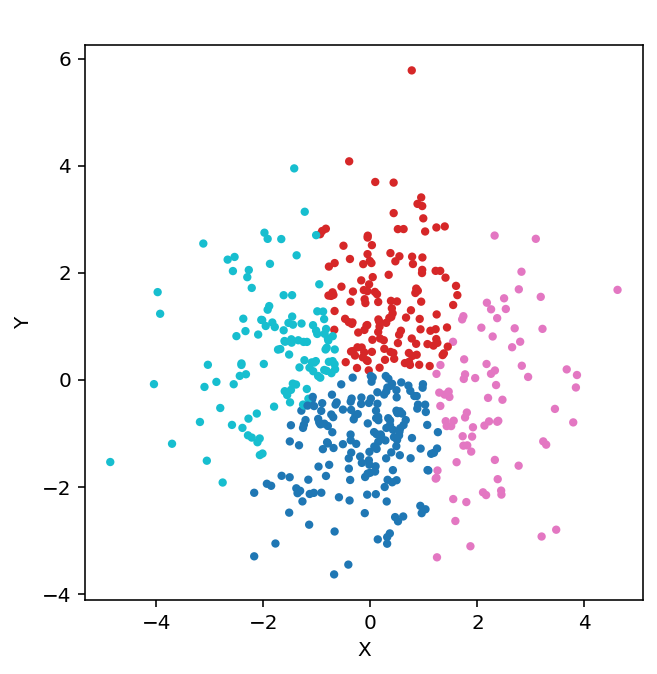}
        \caption{GMM}
        \label{fig:gmm}
    \end{subfigure}

    \begin{subfigure}[b]{0.32\textwidth}
        \includegraphics[width=\textwidth]{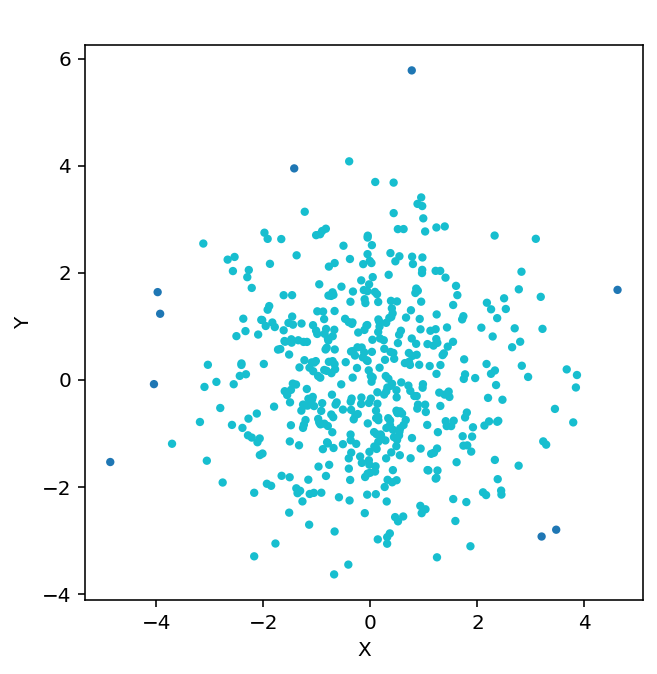}
        \caption{DBSCAN}
        \label{fig:dbscan}
    \end{subfigure}
    \hspace{0.01\textwidth}
    \begin{subfigure}[b]{0.32\textwidth}
        \includegraphics[width=\textwidth]{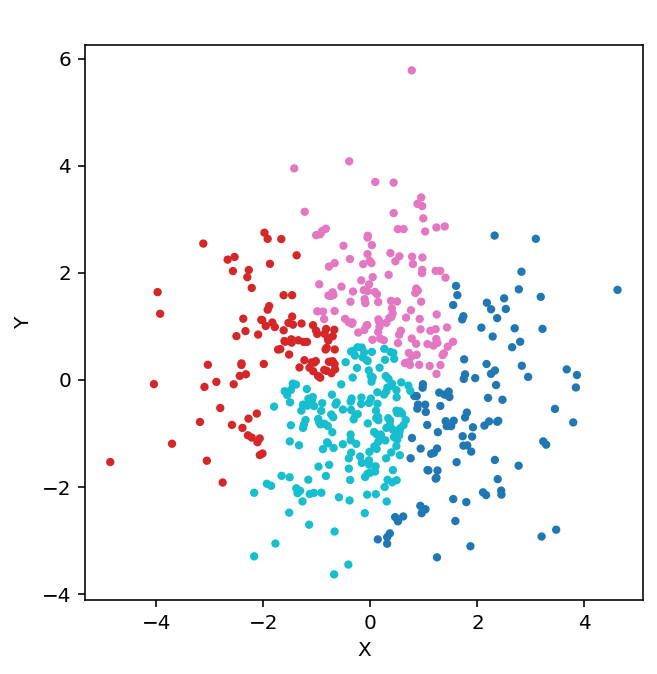}
        \caption{Agglomerative}
        \label{fig:agglo}
    \end{subfigure}
    \hspace{0.01\textwidth}
    \begin{subfigure}[b]{0.32\textwidth}
        \includegraphics[width=\textwidth]{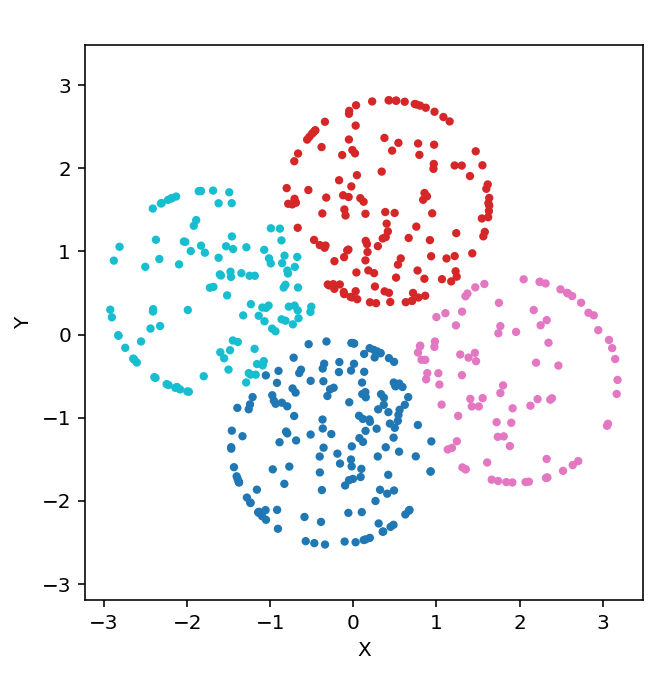}
        \caption{CCC}
        \label{fig:ccc}
    \end{subfigure}
 \end{minipage}
    }
    \caption{Clustering results on the synthetic circular dataset: (a) original scattered data, (b) KMeans, (c) GMM, (d) DBSCAN, (e) Agglomerative, and (f) our proposed CCC method. The visualizations are generated on 500 points with a standard deviation of 1.5 to ensure clear visibility of cluster structures while demonstrating differences in spread and organization across methods.}
    \label{fig:clustering_comparison 500 1.5}
\end{figure*}
This experimental setup enables a clear and direct comparison, ensuring that the observed reductions in entropy under the proposed constrained clustering method can be attributed to its ability to reduce the radial spread of clusters while preserving their angular characteristics. Consequently, the experiments validate the practical utility of the proposed method in achieving structured, interpretable, and compact clustering in scenarios where control over cluster spread is essential. Table \ref{tab:entropy_std_comparison 500} presents the quantitative entropy results following this evaluation strategy.

Figure \ref{fig:clustering_comparison 500 1.5} presents the clustering results on the synthetic circular dataset generated with a standard deviation of 1.5, using five different methods for comparison: KMeans, Gaussian Mixture Models (GMM), DBSCAN, Agglomerative Clustering, and our proposed Constrained Centroid Clustering (CCC).
In the KMeans and GMM plots, where the number of clusters was set to $k$ = $4$, we observe that the dataset is divided into four clusters based on proximity, with each cluster roughly occupying a quadrant-like region. These methods group points by minimizing within-cluster variance but do not explicitly control the maximum spread of clusters, allowing some points to be relatively far from their cluster centers.
The DBSCAN plot shows most points assigned to a single large cluster, with a few outliers scattered around the periphery. Since DBSCAN is density-based, it groups closely packed points while labeling low-density regions as noise. In this experiment, we used an epsilon value of 1.0 and a minimum points threshold of 5. We also experimented with lower epsilon values and varied the minimum points parameter, but due to the relatively uniform density of the synthetic circular dataset, DBSCAN either absorbed outliers into clusters or grouped nearly all points into a single cluster while leaving a few outliers off at the edges. As a result, DBSCAN did not produce distinct, structured clusters in this scenario, unlike KMeans, GMM, or our proposed CCC method.
The Agglomerative Clustering plot shows a result similar to KMeans and GMM, where the data is divided into four clusters based on proximity. However, the exact boundaries of clusters can differ slightly due to the hierarchical linkage criteria used in agglomerative methods.
In the CCC plot, we see a clear structural difference: each cluster forms a neat, circular shape around its center, with no points scattered far from the clusters. This circular arrangement occurs because we place a limit on how far any point can be from its assigned cluster center, pulling in any points that would otherwise lie further away. This constraint not only controls the spread of each cluster but also results in a compact, interpretable structure while preserving the overall shape of the clusters. 
We quantitatively evaluate this structural compactness using the entropy-based measures described earlier, confirming how effectively CCC controls spread and maintains organization within the clusters.

These visual comparisons demonstrate how the proposed CCC method achieves spread-controlled clustering while maintaining interpretability, whereas standard methods allow broader, uncontrolled cluster spreads, and DBSCAN struggles to identify multiple clusters under uniform density conditions. The CCC result visually confirms the method’s design, illustrating how spread constraints translate into clear, bounded clusters in practice.

\textbf{Note:} While we conducted experiments using different standard deviations (std = 1.0, 1.2, 1.5) and larger dataset sizes, we have included visual diagrams in the main text only for the case with 500 points and a standard deviation of 1.5. This choice was made because this configuration provides clear and interpretable cluster structures without excessive overlap, which can occur when using lower standard deviations or larger datasets, making it difficult to distinguish individual points and cluster boundaries in the plots. The results for other configurations, including additional visual diagrams and the corresponding entropy values, are provided in the appendix. This facilitates a clear and systematic exposition, demonstrating that our proposed CCC method remains effective across different data spreads and dataset sizes.

\begin{table*}[t]
\caption{Entropy comparison across clustering methods on synthetic circular data with 500 points under different standard deviations. Lower entropy indicates tighter and more structured clustering, with CCC consistently achieving lower ring and joint entropy while preserving angular structure across varying spreads.}
\centering
\resizebox{\textwidth}{!}{%

\begin{tabular}{lcccc}
\toprule
\textbf{Method} & \textbf{Std} & \textbf{Ring Entropy (bits)} & \textbf{Sector Entropy (bits)} & \textbf{Joint Entropy (bits)} \\
\midrule
KMeans & 1.0 & 1.169 & 2.994 & 4.144 \\
GMM & 1.0 & 1.169 & 2.994 & 4.144 \\
DBSCAN & 1.0 & 1.161 & 2.994 & 4.131 \\
Agglomerative & 1.0 & 1.169 & 2.994 & 4.144 \\
\textbf{CCC} & 1.0 & \textbf{1.017} & \textbf{2.992} & \textbf{3.990} \\
\midrule
KMeans & 1.2 & 1.319 & 2.994 & 4.286 \\
GMM & 1.2 &  1.319 & 2.994 & 4.286 \\
DBSCAN & 1.2 &  1.307 & 2.994 & 4.275 \\
Agglomerative & 1.2&1.319 & 2.994 & 4.286 \\
\textbf{CCC} & 1.2 & \textbf{1.155} & \textbf{2.992} & \textbf{4.074} \\
\midrule
KMeans & 1.5 & 1.411 & 2.994 & 4.391 \\
GMM & 1.5 &  1.411 & 2.994 & 4.391 \\
DBSCAN & 1.5 &  1.400 & 2.993 & 4.380 \\
Agglomerative & 1.5 & 1.411 & 2.994 & 4.391 \\
\textbf{CCC} & 1.5 & \textbf{1.273} & \textbf{2.989} & \textbf{4.182} \\
\bottomrule
\end{tabular}
}

\label{tab:entropy_std_comparison 500}
\end{table*}
\subsection*{Analysis and Insights}
We analyze the clusters using entropy measures to evaluate how effectively CCC reduces spread while preserving angular structure under different standard deviations. Experiments on synthetic circular data with 500 points and varying standard deviations (1.0, 1.2, and 1.5) showed a clear reduction in ring entropy (approximately 10\%) and joint entropy (approximately 5\%) compared to standard methods such as KMeans, GMM, DBSCAN, and Agglomerative Clustering, while maintaining similar sector entropy across all cases. The results of these entropy evaluations are presented in Table~\ref{tab:entropy_std_comparison 500}, illustrating that CCC achieves tighter and more structured clustering while preserving angular distribution consistency under different spreads.

These results demonstrate that our approach effectively reduces radial spread and improves the compactness and structure of clusters without distorting the angular distribution, even under varying data spreads. The findings validate the practical utility of our constrained clustering formulation in achieving structured and interpretable clustering with controlled spread. By reducing radial variability while preserving angular structure, our method is well-suited for applications requiring compact clusters, including sensor network formation, collaborative robotics, and structured pattern analysis in machine learning pipelines. The computational simplicity, interpretability, and scalability of CCC further support its practical deployment in real-world systems.

We note that KMeans, GMM, and Agglomerative clustering exhibit nearly identical entropy results due to the radial symmetry and uniform angular distribution of the synthetic data, combined with the centroid-based nature of these methods, which leads to consistent partitioning under our ring and sector entropy analysis framework.
\paragraph{Note on Dataset Selection:}
As noted in the Introduction, while real-world datasets exhibiting perfect radial symmetry are uncommon, the practical need to control cluster spread remains relevant across applications. However, publicly available datasets with consistent radial and angular structure suitable for ring-sector entropy evaluation are limited. Therefore, we use synthetically generated circular data in this study to provide a controlled, interpretable testbed for validating the proposed CCC method. This choice ensures that the observed improvements in entropy and cluster compactness directly reflect the method's design rather than dataset artifacts. Future work will focus on applying CCC to domain-specific datasets where the radial and angular structure aligns with the proposed evaluation framework.
\section{Conclusion}\label{conclusion}
In this work, we presented a constrained clustering formulation that incorporates a practical spread-control mechanism by limiting the distance between the cluster center and the extremal pattern within each cluster. Building on classical centroid-based clustering, we derived a computationally efficient KKT-based solution that ensures the cluster center remains representative while respecting a user-specified constraint on maximum distance. We demonstrated that the proposed method seamlessly reduces to the unconstrained centroid computation when the constraint is inactive, ensuring consistency with existing clustering approaches.

The proposed constrained clustering approach can be particularly useful in applications requiring bounded spread, including sensor networks, collaborative robotics, and interpretable machine learning systems where resource-aware, structured grouping is essential. Future directions include extending this framework to dynamic data scenarios, exploring adaptive selection of the constraint parameter \(S\) based on application needs, and investigating the integration of the method into larger pipeline tasks such as constrained representation learning and active clustering for practical deployment.
  \bibliographystyle{elsarticle-num} 
\bibliography{ref}

\begin{thebibliography}{10}
\expandafter\ifx\csname url\endcsname\relax
  \def\url#1{\texttt{#1}}\fi
\expandafter\ifx\csname urlprefix\endcsname\relax\def\urlprefix{URL }\fi
\expandafter\ifx\csname href\endcsname\relax
  \def\href#1#2{#2} \def\path#1{#1}\fi

\bibitem{4787647}
O.~Chapelle, B.~Scholkopf, A.~Zien, Eds., Semi-supervised learning (chapelle, o. et al., eds.; 2006) [book reviews], IEEE Transactions on Neural Networks 20~(3) (2009) 542--542.
\newblock \href {https://doi.org/10.1109/TNN.2009.2015974} {\path{doi:10.1109/TNN.2009.2015974}}.

\bibitem{10.5555/2380985}
K.~P. Murphy, Machine Learning: A Probabilistic Perspective, The MIT Press, 2012.

\bibitem{Bishop2006}
C.~M. Bishop, \href{http://www.library.wisc.edu/selectedtocs/bg0137.pdf}{{Pattern Recognition and Machine Learning}}, Vol.~4 of Information science and statistics, Springer, 2006.
\newblock \href {http://arxiv.org/abs/0-387-31073-8} {\path{arXiv:0-387-31073-8}}, \href {https://doi.org/10.1117/1.2819119} {\path{doi:10.1117/1.2819119}}.
\newline\urlprefix\url{http://www.library.wisc.edu/selectedtocs/bg0137.pdf}

\bibitem{10.1145/331499.331504}
A.~K. Jain, M.~N. Murty, P.~J. Flynn, \href{https://doi.org/10.1145/331499.331504}{Data clustering: a review}, ACM Comput. Surv. 31~(3) (1999) 264–323.
\newblock \href {https://doi.org/10.1145/331499.331504} {\path{doi:10.1145/331499.331504}}.
\newline\urlprefix\url{https://doi.org/10.1145/331499.331504}

\bibitem{1427769}
R.~Xu, D.~Wunsch, Survey of clustering algorithms, IEEE Transactions on Neural Networks 16~(3) (2005) 645--678.
\newblock \href {https://doi.org/10.1109/TNN.2005.845141} {\path{doi:10.1109/TNN.2005.845141}}.

\bibitem{10.1007/11564126_11}
I.~Davidson, S.~S. Ravi, Agglomerative hierarchical clustering with constraints: Theoretical and empirical results, in: A.~M. Jorge, L.~Torgo, P.~Brazdil, R.~Camacho, J.~Gama (Eds.), Knowledge Discovery in Databases: PKDD 2005, Springer Berlin Heidelberg, Berlin, Heidelberg, 2005, pp. 59--70.

\bibitem{5206852}
Z.~Li, J.~Liu, X.~Tang, Constrained clustering via spectral regularization, in: 2009 IEEE Conference on Computer Vision and Pattern Recognition, 2009, pp. 421--428.
\newblock \href {https://doi.org/10.1109/CVPR.2009.5206852} {\path{doi:10.1109/CVPR.2009.5206852}}.

\bibitem{8412085}
E.~Min, X.~Guo, Q.~Liu, G.~Zhang, J.~Cui, J.~Long, A survey of clustering with deep learning: From the perspective of network architecture, IEEE Access 6 (2018) 39501--39514.
\newblock \href {https://doi.org/10.1109/ACCESS.2018.2855437} {\path{doi:10.1109/ACCESS.2018.2855437}}.

\bibitem{bachem2017practicalcoresetconstructionsmachine}
O.~Bachem, M.~Lucic, A.~Krause, \href{https://arxiv.org/abs/1703.06476}{Practical coreset constructions for machine learning} (2017).
\newblock \href {http://arxiv.org/abs/1703.06476} {\path{arXiv:1703.06476}}.
\newline\urlprefix\url{https://arxiv.org/abs/1703.06476}

\bibitem{bradley2000constrained}
P.~S. Bradley, K.~P. Bennett, A.~Demiriz, Constrained k-means clustering, Microsoft Research, Redmond 20~(0) (2000) 0.

\bibitem{JMLR:v11:vinh10a}
N.~X. Vinh, J.~Epps, J.~Bailey, \href{http://jmlr.org/papers/v11/vinh10a.html}{Information theoretic measures for clusterings comparison: Variants, properties, normalization and correction for chance}, Journal of Machine Learning Research 11~(95) (2010) 2837--2854.
\newline\urlprefix\url{http://jmlr.org/papers/v11/vinh10a.html}

\bibitem{MEILA2007873}
M.~Meilă, \href{https://www.sciencedirect.com/science/article/pii/S0047259X06002016}{Comparing clusterings—an information based distance}, Journal of Multivariate Analysis 98~(5) (2007) 873--895.
\newblock \href {https://doi.org/https://doi.org/10.1016/j.jmva.2006.11.013} {\path{doi:https://doi.org/10.1016/j.jmva.2006.11.013}}.
\newline\urlprefix\url{https://www.sciencedirect.com/science/article/pii/S0047259X06002016}

\bibitem{mcqueen1967some}
J.~B. McQueen, Some methods of classification and analysis of multivariate observations, in: Proc. of 5th Berkeley Symposium on Math. Stat. and Prob., 1967, pp. 281--297.

\bibitem{wagstaff2001constrained}
K.~Wagstaff, C.~Cardie, S.~Rogers, S.~Schr{\"o}dl, et~al., Constrained k-means clustering with background knowledge, in: Icml, Vol.~1, 2001, pp. 577--584.

\bibitem{ester1996density}
M.~Ester, H.-P. Kriegel, J.~Sander, X.~Xu, et~al., A density-based algorithm for discovering clusters in large spatial databases with noise, in: kdd, Vol.~96, 1996, pp. 226--231.

\bibitem{lin2002divergence}
J.~Lin, Divergence measures based on the shannon entropy, IEEE Transactions on Information theory 37~(1) (2002) 145--151.

\bibitem{zhong2003unified}
S.~Zhong, J.~Ghosh, A unified framework for model-based clustering, Journal of machine learning research 4~(Nov) (2003) 1001--1037.

\end{thebibliography}
\section*{Appendix:}
\subsection*{Detailed PMF and Entropy Calculations}
We provide the detailed computation of ring, sector, and joint PMFs and their corresponding entropy values using the experimental counts for the example illustrated in Figure~\ref{fig:ring-sector}.
\subsection*{Ring PMF}
Given:
\[
\tilde{n}_1 = 8, \quad \tilde{n}_2 = 10, \quad \tilde{n}_3 = 10, \quad T = 28
\]

The ring PMFs are:
\[
p_1 = \frac{8}{28} = 0.286, \quad
p_2 = \frac{10}{28} = 0.357, \quad
p_3 = \frac{10}{28} = 0.357
\]

The ring entropy:
\[
H(P) = - \sum_{i=1}^{3} p_i \log_2 p_i
\]
\[
= -0.286 \log_2 0.286 - 0.357 \log_2 0.357 - 0.357 \log_2 0.357
\]
\[
= 0.517 + 0.530 + 0.530 = 1.577 \text{ bits}
\]
\subsection*{Sector PMF}
Given:
\[
\Gamma_1 = 11, \quad \Gamma_2 = 6, \quad \Gamma_3 = 5, \quad \Gamma_4 = 6, \quad T = 28
\]
The sector PMFs are:
\[
\begin{aligned}
S_1 &= \frac{11}{28} = 0.393,\quad
S_2 = \frac{6}{28} = 0.214, \\
S_3 &= \frac{5}{28} = 0.179,\quad
S_4 = \frac{6}{28} = 0.214
\end{aligned}
\]
The sector entropy is:
\begin{IEEEeqnarray}{rCl}
H(S) &=& - \sum_{j=1}^{4} S_j \log_2 S_j \nonumber\\
&=& -0.393 \log_2 0.393
      - 0.214 \log_2 0.214 \nonumber\\
&&\quad - 0.179 \log_2 0.179
      - 0.214 \log_2 0.214 \nonumber\\
&=& 0.531 + 0.476 + 0.445 + 0.476 \nonumber\\
&=& 1.928 \text{ bits} \IEEEyesnumber
\end{IEEEeqnarray}
\subsection*{Joint PMF}
Given the distribution:
\[
\begin{bmatrix}
3 & 2 & 1 & 2 \\
4 & 2 & 2 & 2 \\
4 & 2 & 2 & 2
\end{bmatrix}
\]
The joint PMFs:
\[
p_{i,j} = \frac{n_{i,j}}{28}
\]
The joint entropy:
\[
H(P_{i,j}) = - \sum_{i=1}^{3} \sum_{j=1}^{4} p_{i,j} \log_2 p_{i,j}
\]
\begin{equation}
\resizebox{\linewidth}{!}{$
\begin{aligned}
H(P_{i,j}) &= - \sum_{i=1}^{3} \sum_{j=1}^{4} p_{i,j} \log_2 p_{i,j} \\
&= - \Big[ 
p_{1,1} \log_2 p_{1,1} + \cdots + p_{3,4} \log_2 p_{3,4} 
\Big] \\
&= - \Big[ 
\frac{3}{28} \log_2 \frac{3}{28} + \cdots + \frac{2}{28} \log_2 \frac{2}{28}
\Big] \\
&= - \Big[
0.1071 \times (-3.23) + 0.0714 \times (-3.81) + 0.0357 \times (-4.81) \\
&\quad + 0.0714 \times (-3.81) + 0.1429 \times (-2.81) + 0.0714 \times (-3.81) \\
&\quad + 0.0714 \times (-3.81) + 0.0714 \times (-3.81) + 0.1429 \times (-2.81) \\
&\quad + 0.0714 \times (-3.81) + 0.0714 \times (-3.81) + 0.0714 \times (-3.81)
\Big] \\
&= 0.346 + 0.272 + 0.172 + 0.272 + 0.402 + 0.272 + 0.272 \\
&\quad + 0.272 + 0.402 + 0.272 + 0.272 + 0.272 \\
&= 3.498 \text{ bits}
\end{aligned}
$}
\end{equation}

\section*{Additional Experimental Results}
\begin{figure*}[t]
     \centering
    \resizebox{0.82\textwidth}{!}{ 
    \begin{minipage}{\textwidth}

    \begin{subfigure}[b]{0.32\textwidth}
        \includegraphics[width=\textwidth]{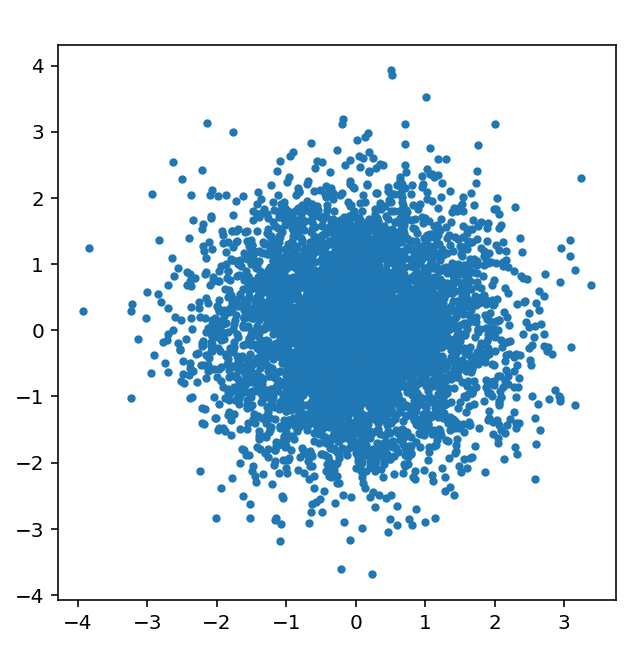}
        \caption{Original data}
        \label{fig:original}
    \end{subfigure}
    \hspace{0.01\textwidth}
    \begin{subfigure}[b]{0.32\textwidth}
        \includegraphics[width=\textwidth]{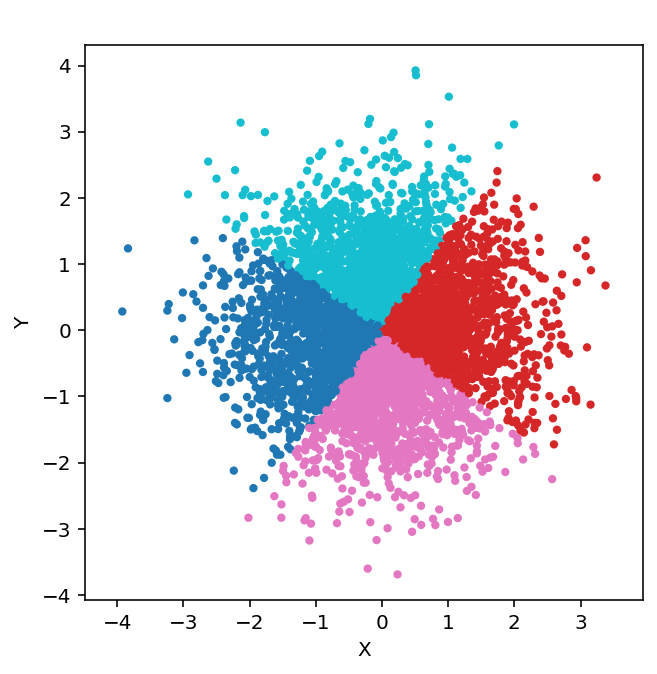}
        \caption{KMeans}
        \label{fig:kmeans}
    \end{subfigure}
    \hspace{0.01\textwidth}
    \begin{subfigure}[b]{0.32\textwidth}
        \includegraphics[width=\textwidth]{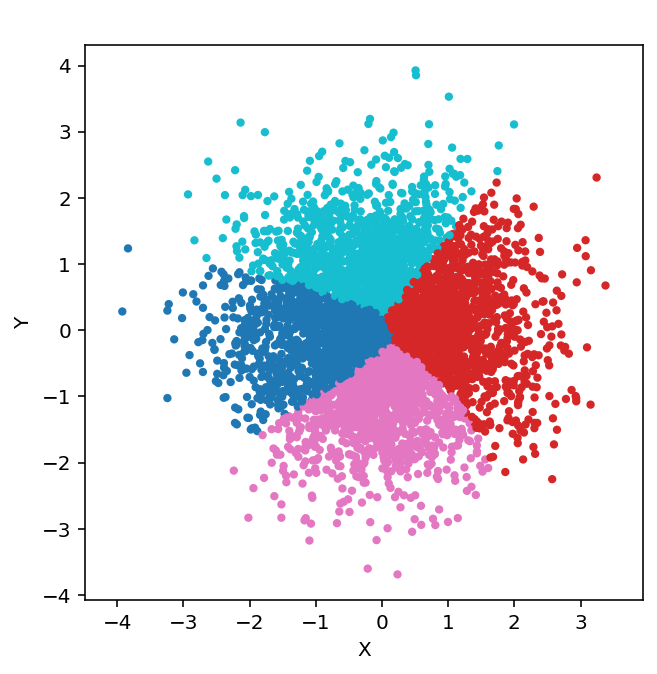}
        \caption{GMM}
        \label{fig:gmm}
    \end{subfigure}

    \begin{subfigure}[b]{0.32\textwidth}
        \includegraphics[width=\textwidth]{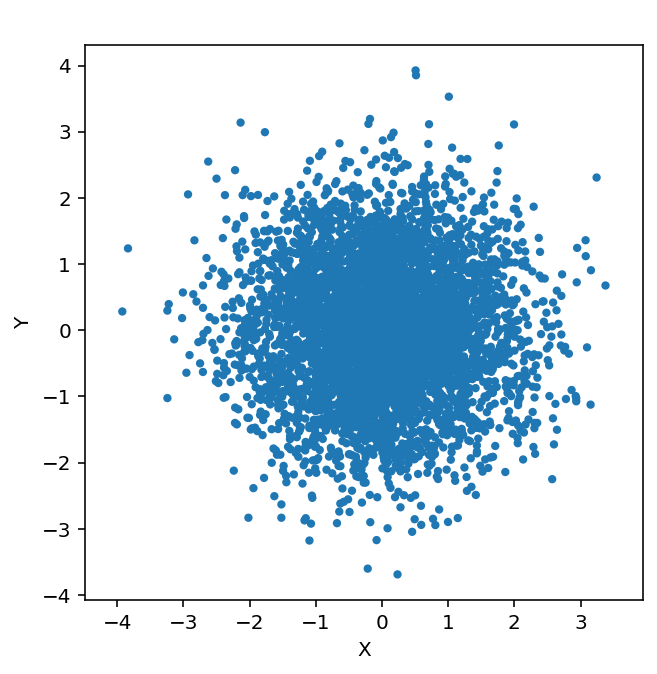}
        \caption{DBSCAN}
        \label{fig:dbscan}
    \end{subfigure}
    \hspace{0.01\textwidth}
    \begin{subfigure}[b]{0.32\textwidth}
        \includegraphics[width=\textwidth]{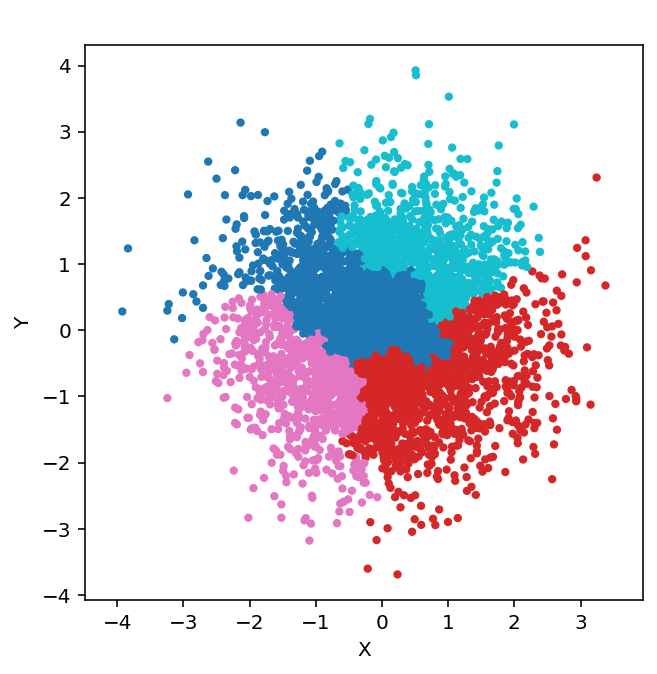}
        \caption{Agglomerative}
        \label{fig:agglo}
    \end{subfigure}
    \hspace{0.01\textwidth}
    \begin{subfigure}[b]{0.32\textwidth}
        \includegraphics[width=\textwidth]{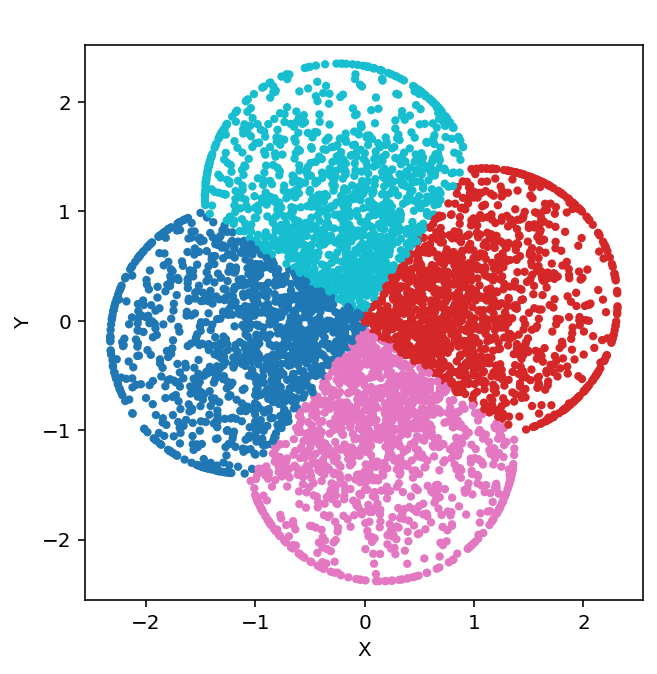}
        \caption{CCC}
        \label{fig:ccc}
    \end{subfigure}
 \end{minipage}
    }
    \caption{Clustering results on the synthetic circular dataset consisting of 5000 points with a standard deviation of 1: (a) original scattered data, (b) KMeans, (c) GMM, (d) DBSCAN, (e) Agglomerative, and (f) our proposed CCC method.}
    \label{fig:clustering_comparison 5000 1std}
\end{figure*}

\begin{figure*}[t]
    \centering
    \resizebox{0.82\textwidth}{!}{ 
    \begin{minipage}{\textwidth}
    \begin{subfigure}[b]{0.32\textwidth}
        \includegraphics[width=\textwidth]{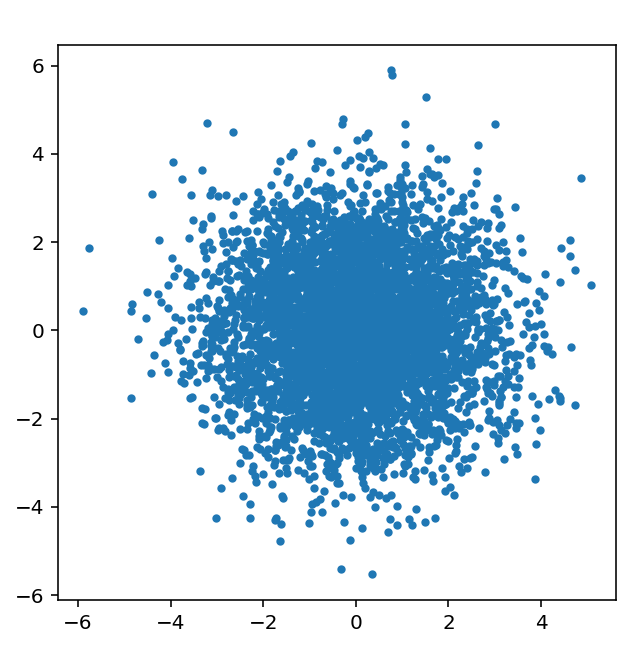}
        \caption{Original Data}
        \label{fig:original}
    \end{subfigure}
    \hspace{0.01\textwidth}
    \begin{subfigure}[b]{0.32\textwidth}
        \includegraphics[width=\textwidth]{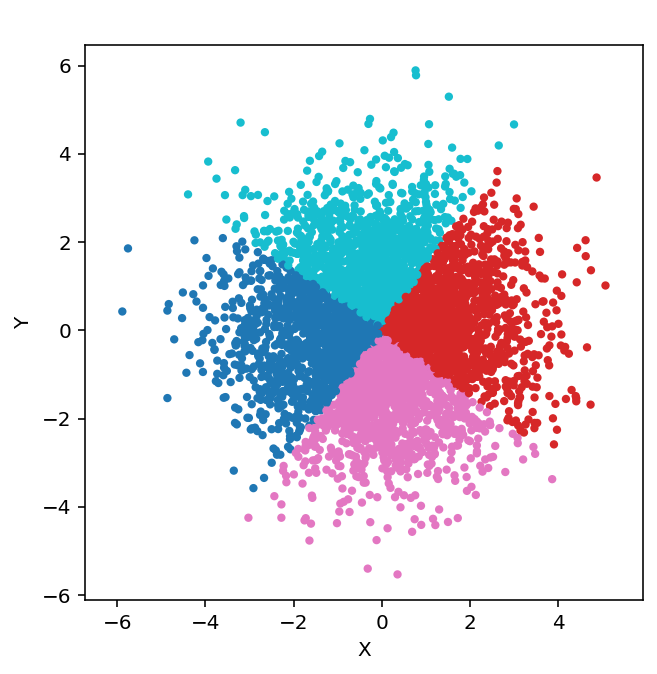}
        \caption{KMeans}
        \label{fig:kmeans}
    \end{subfigure}
    \hspace{0.01\textwidth}
    \begin{subfigure}[b]{0.32\textwidth}
        \includegraphics[width=\textwidth]{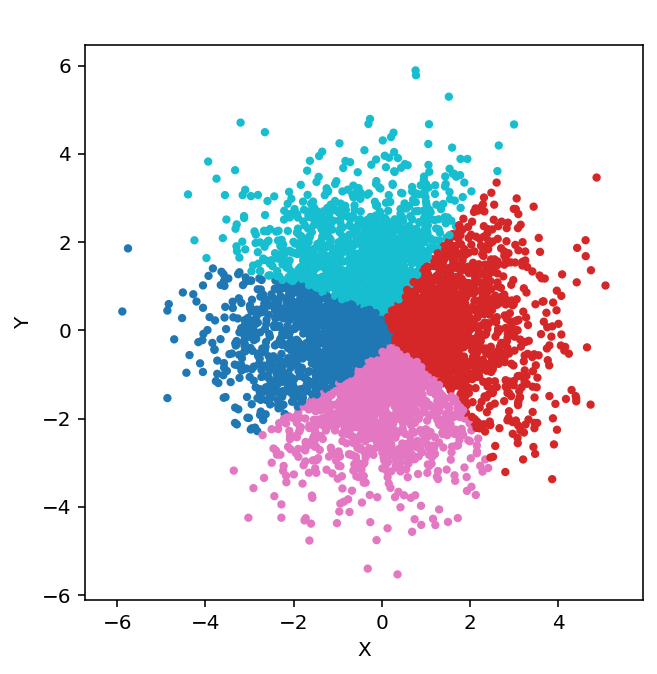}
        \caption{GMM}
        \label{fig:gmm}
    \end{subfigure}

    \begin{subfigure}[b]{0.32\textwidth}
        \includegraphics[width=\textwidth]{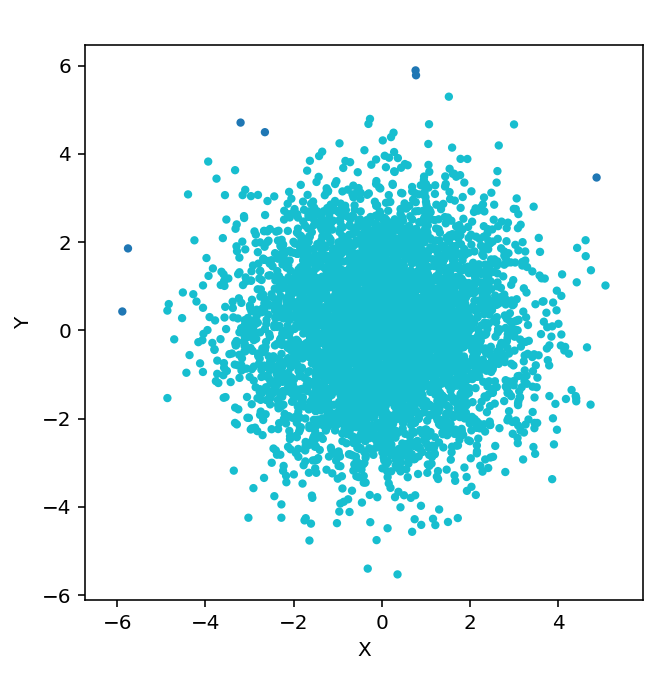}
        \caption{DBSCAN}
        \label{fig:dbscan}
    \end{subfigure}
    \hspace{0.01\textwidth}
    \begin{subfigure}[b]{0.32\textwidth}
        \includegraphics[width=\textwidth]{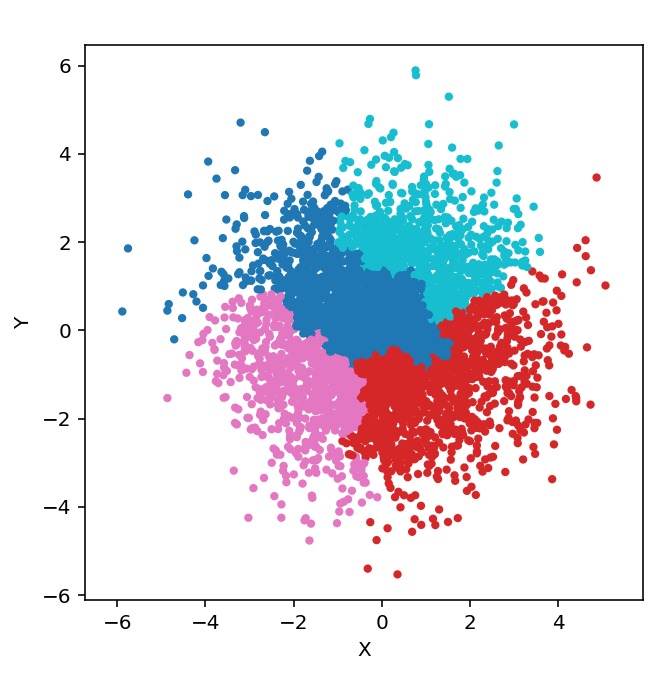}
        \caption{Agglomerative}
        \label{fig:agglo}
    \end{subfigure}
    \hspace{0.01\textwidth}
    \begin{subfigure}[b]{0.32\textwidth}
        \includegraphics[width=\textwidth]{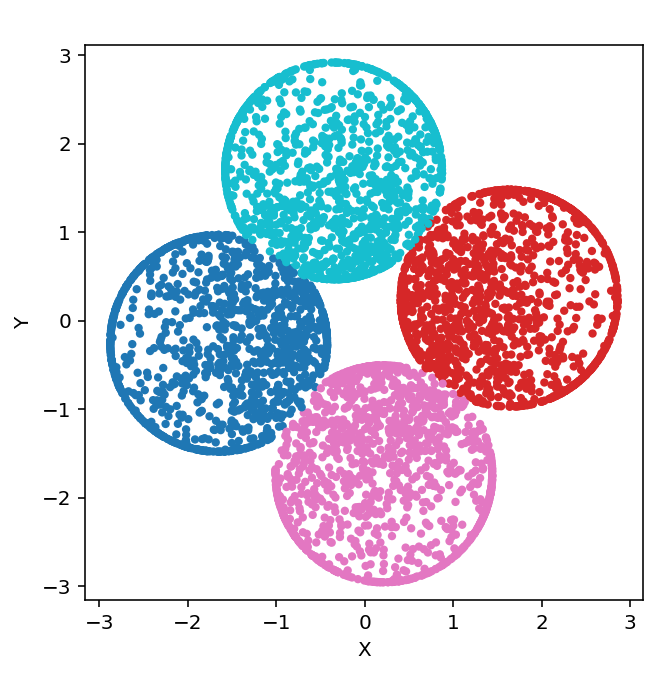}
        \caption{CCC}
        \label{fig:ccc}
    \end{subfigure}
 \end{minipage}
    }
    \caption{Clustering results on the synthetic circular dataset consisting of 5000 points with a standard deviation of 1.5: (a) original scattered data, (b) KMeans, (c) GMM, (d) DBSCAN, (e) Agglomerative, and (f) our proposed CCC method.}
    \label{fig:clustering_comparison 5000 1.5std}
\end{figure*}

\begin{figure*}[t]
    \centering
     \resizebox{0.82\textwidth}{!}{ 
    \begin{minipage}{\textwidth}
    \begin{subfigure}[b]{0.32\textwidth}
        \includegraphics[width=\textwidth]{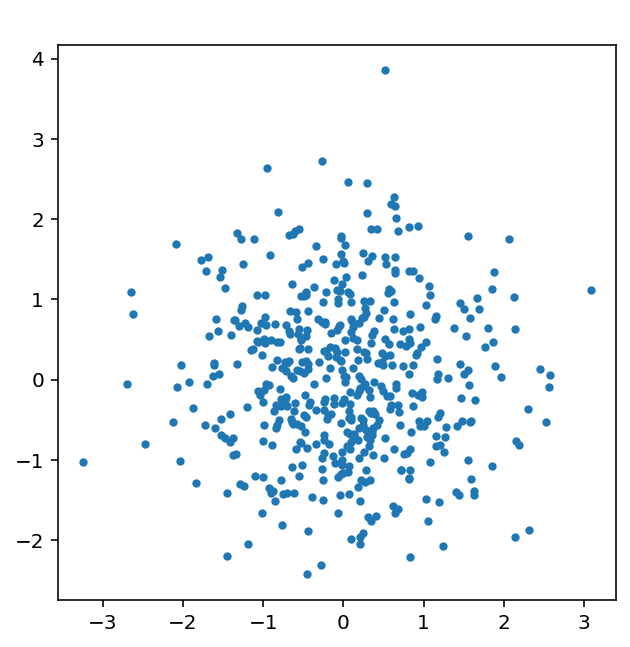}
        \caption{Original Data}
        \label{fig:original}
    \end{subfigure}
    \hspace{0.01\textwidth}
    \begin{subfigure}[b]{0.32\textwidth}
        \includegraphics[width=\textwidth]{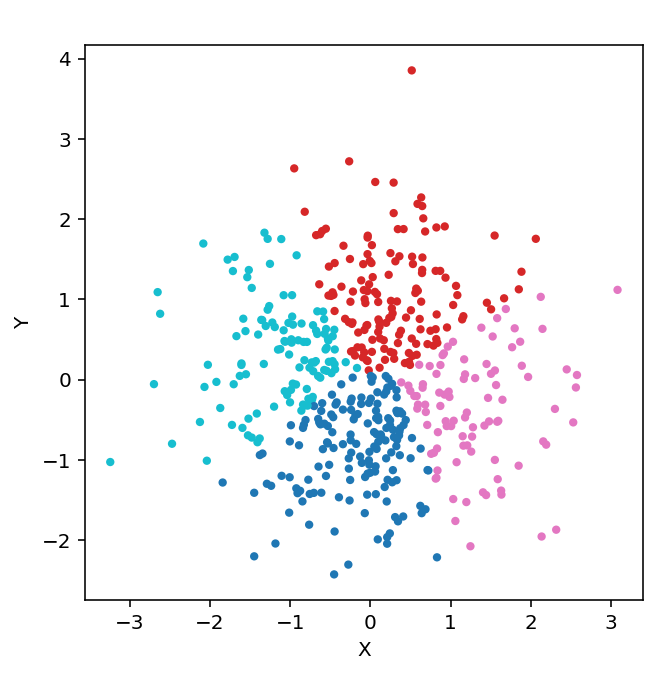}
        \caption{KMeans}
        \label{fig:kmeans}
    \end{subfigure}
    \hspace{0.01\textwidth}
    \begin{subfigure}[b]{0.32\textwidth}
        \includegraphics[width=\textwidth]{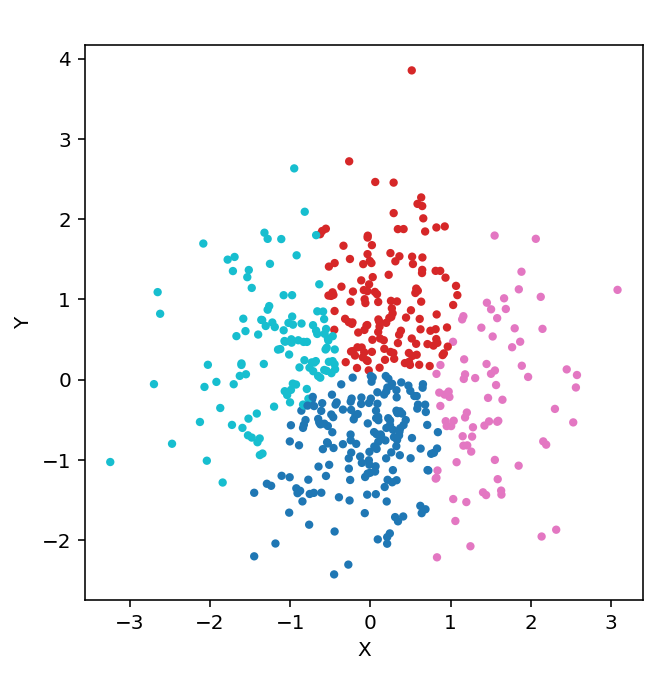}
        \caption{GMM}
        \label{fig:gmm}
    \end{subfigure}

    \begin{subfigure}[b]{0.32\textwidth}
        \includegraphics[width=\textwidth]{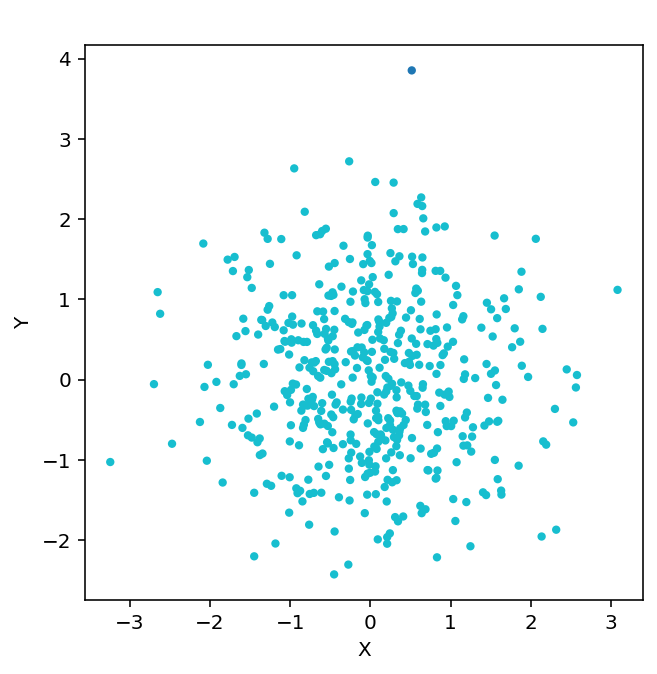}
        \caption{DBSCAN}
        \label{fig:dbscan}
    \end{subfigure}
    \hspace{0.01\textwidth}
    \begin{subfigure}[b]{0.32\textwidth}
        \includegraphics[width=\textwidth]{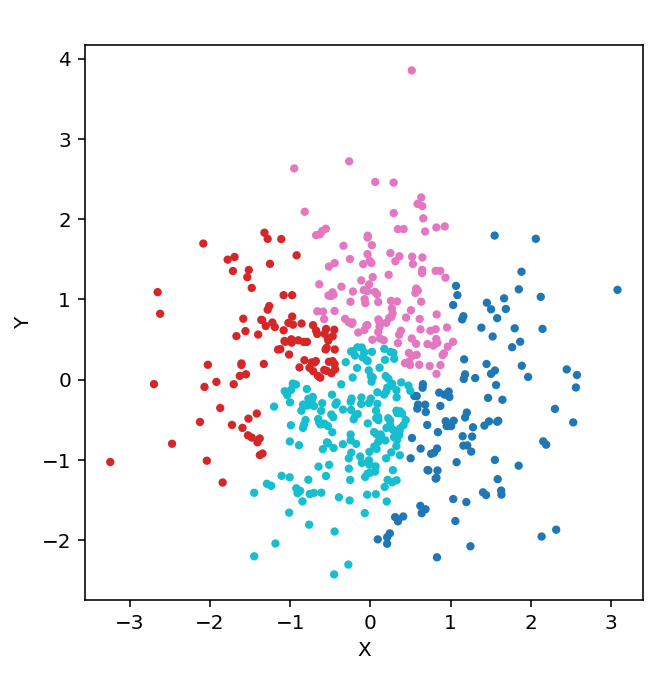}
        \caption{Agglomerative}
        \label{fig:agglo}
    \end{subfigure}
    \hspace{0.01\textwidth}
    \begin{subfigure}[b]{0.32\textwidth}
        \includegraphics[width=\textwidth]{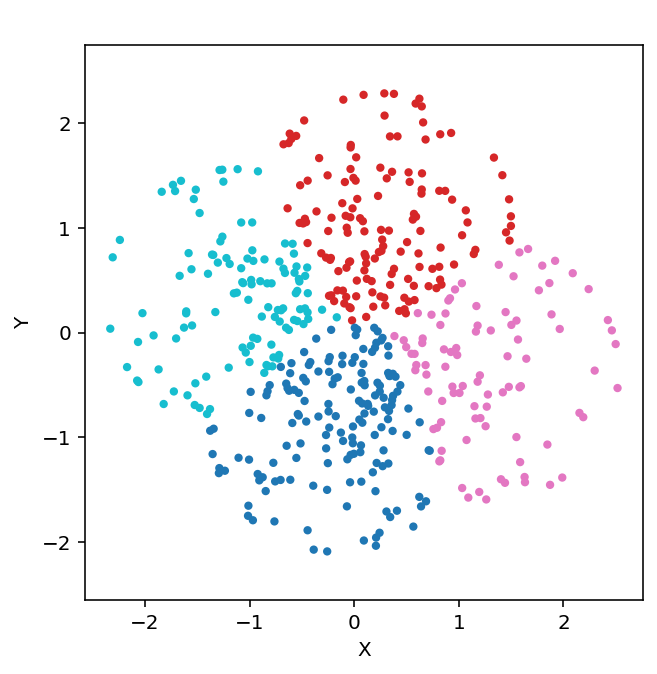}
        \caption{CCC}
        \label{fig:ccc}
    \end{subfigure}
 \end{minipage}
    }
    \caption{Clustering results on the synthetic circular dataset consisting of 500 points with a standard deviation of 1: (a) original scattered data, (b) KMeans, (c) GMM, (d) DBSCAN, (e) Agglomerative, and (f) our proposed CCC method.}
    \label{fig:clustering_comparison 500 1 std}
\end{figure*}

Though we conducted experiments with other dataset sizes and spread conditions to comprehensively evaluate the methods, we include these results here primarily for illustration and completeness. Specifically, we provide additional experimental diagrams and tabulated entropy evaluations under varied dataset sizes and spread conditions to complement the main results. Figures \ref{fig:clustering_comparison 5000 1std} and \ref{fig:clustering_comparison 5000 1.5std} show visualizations for 5000 data points with standard deviations of 1.0 and 1.5, respectively, illustrating how the clustering structure evolves with data density and spread. Figure \ref{fig:clustering_comparison 500 1 std} presents a visualization for 500 data points with a standard deviation of 1.0, while results for 500 points with a standard deviation of 1.5 are included in the main paper for direct comparison. Additionally, Table \ref{tab:entropy_std_comparison 5000} provides entropy values for 5000 data points under different standard deviations (1.0, 1.2, 1.5), enabling quantitative comparison across methods under higher data densities. These additional results offer a broader perspective on the stability and effectiveness of the proposed CCC method and other baseline methods across different dataset scales and spread conditions, supporting the observations discussed in the main paper.
\begin{table*}[t]
\caption{Entropy comparison across clustering methods on synthetic circular data with 5000 points under different standard deviations. Lower entropy indicates tighter and more structured clustering, with CCC consistently achieving lower ring and joint entropy while preserving angular structure across varying spreads.}
\centering
\resizebox{\textwidth}{!}{%

\begin{tabular}{lcccc}
\toprule
\textbf{Method} & \textbf{Std} & \textbf{Ring Entropy (bits)} & \textbf{Sector Entropy (bits)} & \textbf{Joint Entropy (bits)} \\
\midrule
KMeans & 1.0 & 1.203 & 3.000 & 4.201 \\
GMM & 1.0 & 1.203 & 3.000 & 4.201 \\
DBSCAN & 1.0 & 1.203 & 3.000 & 4.201 \\
Agglomerative & 1.0 & 1.203 & 3.000 & 4.201\\
\textbf{CCC} & 1.0 & \textbf{0.965} & \textbf{2.999} & \textbf{3.964} \\
\midrule
KMeans & 1.2 & 1.323 & 3.000 & 4.320 \\
GMM & 1.2 &  1.323 & 3.000 & 4.320 \\
DBSCAN & 1.2 &  1.322 & 3.000 & 4.319 \\
Agglomerative & 1.2 &  1.323 & 3.000 & 4.320 \\
\textbf{CCC} & 1.2 & \textbf{1.091} & \textbf{2.998} & \textbf{4.073} \\
\midrule
KMeans & 1.5 & 1.438 & 3.000 & 4.436 \\
GMM & 1.5 &  1.438 & 3.000 & 4.436 \\
DBSCAN & 1.5 &  1.437 & 3.000 & 4.435 \\
Agglomerative & 1.5 & 1.438 & 3.000 & 4.436 \\
\textbf{CCC} & 1.5 & \textbf{1.356} & \textbf{2.991} & \textbf{4.336} \\
\bottomrule
\end{tabular}
}

\label{tab:entropy_std_comparison 5000}
\end{table*}
\begin{IEEEbiography}[{\includegraphics[width=1in,height=1.25in,clip,keepaspectratio]{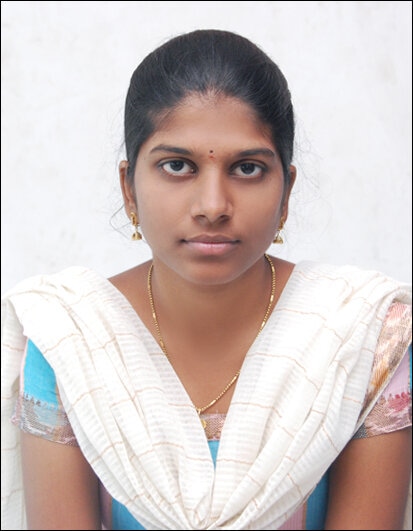}}]{Sowmini Veeramachaneni}
received the B.Tech. degree in Computer Science and Engineering from Jawaharlal Nehru Technological University Kakinada (JNTUK), India, and the M.Tech. and Ph.D. degrees in Computer Science from the University of Hyderabad, India. She qualified the CSIR-UGC-NET Junior Research Fellowship (JRF). She is currently an Assistant Professor in the Department of Artificial Intelligence and Computer Science at Mahindra University, Hyderabad, India. Her research interests include recommender systems, machine learning, and optimization techniques. She is a member of the IEEE and the ACM.
\end{IEEEbiography}

\begin{IEEEbiography}[{\includegraphics[width=1in,height=1.25in,clip,keepaspectratio]{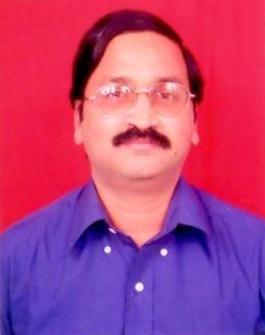}}]{Ramamurthy Garimella}{(Senior Member,
IEEE) received the B.Tech. degree in electronics
and communication engineering from SV University, Tirupati, India, in 1984, the M.S. degree in
electrical engineering from Louisiana State University, USA, in 1986, and the Ph.D. degree in
computer engineering from Purdue University,
West Lafayette, IN, USA, in 1989. Currently, he is
a Professor of Artificial Intelligence and Computer Science at Mahindra Ecole Centrale, Hyderabad, India.
He has around 30 years of teaching and industrial experience. He has around 300 research papers to his credit. He is a Senior Member of ACM and a fellow of IETE, India. His areas of research interests include wireless sensor networks, neural and fuzzy logic, control systems, artificial intelligence, machine learning, the IoT, and signal processing. He received many awards,
including the Rashtriya Gaurav Award.}

\end{IEEEbiography}

\end{document}